\documentclass[10pt,twocolumn,letterpaper]{article}
\usepackage{cvpr} 
\makeatletter
\@namedef
{ver@everyshi.sty}{}
\makeatother
\usepackage{graphicx}
\usepackage{amsmath}
\usepackage{amssymb}
\usepackage{booktabs}
\usepackage{bm}
\usepackage{tabularx, makecell}

\usepackage[pagebackref,breaklinks,colorlinks,bookmarks=false]{hyperref}
		
\usepackage[capitalize]{cleveref}
\crefname{section}{Sec.}{Secs.}
\Crefname{section}{Section}{Sections}
\Crefname{table}{Table}{Tables}
\crefname{table}{Tab.}{Tabs.}
\usepackage{tikz}
\usepackage{times}
\usepackage{epsfig}
\usepackage{graphicx}
\usepackage{amsmath}
\usepackage{amssymb}
\usepackage[linesnumbered,ruled,lined]{algorithm2e}
\usepackage{forest}
\usetikzlibrary{arrows.meta}
\usepackage{comment} 
\usepackage{adjustbox}
\usepackage{multirow}

\usepackage[pagebackref,breaklinks,colorlinks]{hyperref}
\usepackage{xcolor}

\def\cvprPaperID{****} 
\def\confName{ICCV}
\def\confYear{2021}
\graphicspath{{./figs/}{./figs/result/}}

\tikzset{
  every leaf node/.style={draw=red,rectangle,align=center},
  every tree node/.style={draw=black,rectangle, align=center},
  tt/.style={font=\ttfamily},
}
\forestset{tikzQtree/.style={for tree={if n children=0{
        node options=every leaf node/.try}{node options=every tree node/.try}}}}
\begin{document}

\title{LEURN: Learning Explainable Univariate Rules with Neural Networks}

\author{Caglar Aytekin\\
{\tt\small cagosmail@gmail.com}
}

\maketitle

\begin{abstract}
   In this paper, we propose LEURN: a neural network architecture that learns univariate decision rules.
   LEURN is a white-box algorithm that results into univariate trees and makes explainable decisions in every stage. 
   In each layer, LEURN finds a set of univariate rules based on an embedding of the previously checked rules and their corresponding responses.
   Both rule finding and final decision mechanisms are weighted linear combinations of these embeddings, hence contribution of all rules are clearly formulated and explainable.
   LEURN can select features, extract feature importance, provide semantic similarity between a pair of samples, be used in a generative manner and can give a confidence score.
   Thanks to a smoothness parameter, LEURN can also controllably behave like decision trees or vanilla neural networks.
   Besides these advantages, LEURN achieves comparable performance to state-of-the-art methods across 30 tabular datasets for classification and regression problems.
\end{abstract}

\section{Introduction}

Although there is an immense amount of work on explainable artificial intelligence, a human explainable white box neural network still does not exist.
The efforts in explainable artificial intelligence has been focused on saliency maps \cite{simonyan2013deep},\cite{zeiler2014visualizing},\cite{zhou2016learning}, \cite{selvaraju2017grad}, \cite{chattopadhay2018grad}, \cite{draelos2020use},\cite{zhang2018top}, \cite{muhammad2020eigen}, \cite{collins2018deep}, approximation by explainable methods \cite{humbird2018deep},\cite{yang2018deep,kontschieder2015deep,sethi1990entropy},\cite{frosst2017distilling},\cite{wu2018beyond}, or hybrid models \cite{mullapudi2018hydranets},\cite{redmon2017yolo9000},\cite{murdock2016blockout},\cite{murthy2016deep},\cite{roy2016monocular},\cite{ahmed2016network},\cite{mcgill2017deciding}, \cite{veit2018convolutional},\cite{wan2020nbdt}.
Many of these works employ some form of decision trees in order to reach explainability.
This is due to the commonly assumed understanding that decision trees are more explainable than neural networks since they extract clear rules on input features.

An interesting fact that have been pointed out by several studies, is that neural networks can be equivalently represented as decision trees \cite{aytekin2022neural}, \cite{zhang2018tropical}, \cite{balestriero2017neural}, \cite{nguyen2020towards}, \cite{sudjianto2020unwrapping}.
This seems to conflict with trying to explain neural networks with decision trees, as they are decision trees themselves.
However, the nature of the decision trees extracted by neural networks are different from commonly used ones in two aspects: they are multivariate and their branches grows exponentially with depth.
Exponentially growing number of branches hurts global explainability as it may even become infeasible to store the tree.
Although one may still focus on local (sample-wise) explanations, multivariate rules are much harder to explain than univariate ones, as they mix features.
Finally, with increasing neural network depth, it becomes even harder to make sense of these rules as the contribution of each rule is not clear.

Motivated by above observations, in this paper, we propose a special neural network architecture (LEURN) that provides an exact univariate decision tree where each rule contributes linearly to the next rule selection and the final decision.
In the first block, LEURN directly learns univariate rules and outputs an embedding of the rules and response to the rules.
Then a linear layer is applied on this embedding in order to find the next rule.
In successive blocks, all embeddings from previous blocks are concatenated and given as input to a linear layer.
Final layer takes all previous embeddings as input, consists of a single linear layer and an activation based on the application -sigmoid for binary classification, softmax for multiclass classification and identity for regression.
Thus, there is a white-box rule selection and final decision making processes with clear contributions of previous rules in a linear manner.
Besides human explainability, the proposed architecture provides additional properties such as: feature importance, feature selection, pair-wise semantic similarity, generative usage, and confidence scoring. 

\section{Related Work}

\subsection{Neural Networks for Tabular Data}

We have reviewed the difference of commonly used decision trees and neural network extracted ones in the previous chapter.
This difference also shows itself in performance of neural networks and decision trees in tabular data. 
Particularly, in \cite{shwartz2022tabular} , \cite{borisov2022deep} and \cite{grinsztajn2022tree}, extensive comparisons of deep learning methods and tree-based methods were made. 
The comparison output tends to be in favor of tree-based methods in terms of both accuracy, training speed and robustness to dataset size.
Following our discussion in previous section, the performance difference should come either from the multivariate rules or the exponentially growing branches, as these are the only differences between neural network trees and common ones.
This highly motivates us to evaluate LEURN on tabular data, as it removes one of these possibilities since it results into univariate trees.
Hence, we believe the results of this evaluation is crucial to understand whether the performance gap comes from exponentially growing branches.

Motivated by above, our application focus is structured data, hence we review deep learning based methods for tabular data next. 

Majority of works investigate feature transformation \cite{gorishniy2022embeddings}\cite{sun2019supertml}\cite{zhu2021converting}\cite{buturovic2020novel}, transfer-learning \cite{levin2022transfer}\cite{rubachev2022revisiting}, self-supervised learning \cite{bahri2021scarf}\cite{arik2021tabnet}\cite{huang2020tabtransformer},\cite{somepalli2021saint}, attention \cite{arik2021tabnet}\cite{huang2020tabtransformer}\cite{somepalli2021saint}, regularizations \cite{kadra2021well}, univariate decisions \cite{agarwal2021neural}\cite{radenovic2022neural}  or explicitly tree-like models \cite{joseph2022gate}\cite{popov2019neural} in order to improve deep learning performance on tabular data. 

\textbf{Feature Transformation:}
The approach in \cite{sun2019supertml} suggests transforming tabular data into two-dimensional embeddings, which can be thought of as an image-like format. 
These embeddings are then utilized as the input for traditional convolutional neural networks.
\cite{zhu2021converting} and \cite{buturovic2020novel} also takes a similar approach.
\cite{gorishniy2022embeddings} finds embedding on numerical features are beneficial for deep learning methods.

\textbf{Transfer Learning:}
In \cite{levin2022transfer}, the authors highlight a key advantage of neural models over others: their ability to learn reusable features and to be fine-tuned in new domains. The authors show this via using upstream data.
\cite{rubachev2022revisiting} also investigates pre-training strategies for tabular data.

\textbf{Attention:}
\cite{arik2021tabnet} uses sequential attention to select relevant features at each decision step.
Tabtransformer \cite{huang2020tabtransformer} also makes use of self-attention.
SAINT \cite{somepalli2021saint} applies attention differently: Both on rows (samples) and columns (features).
\cite{kossen2021self} is another method that utilizes both row and column attention.

\textbf{Self Supervised Learning:}
\cite{arik2021tabnet} \cite{huang2020tabtransformer}\cite{somepalli2021saint}\cite{bahri2021scarf} provide ways to incorporate self-supervised learning with unlabeled data to achieve to better performance especially in small dataset size regime.

\textbf{Regularization:}
\cite{kadra2021well} investigates the effects of regularization combinations on MLPs' tabular data performance and finds that with correct regularization, MLPs can outperform many alternatives.

\textbf{Univariate Decisions:}
The Neural Additive Models (NAMs) \cite{agarwal2021neural} are a type of ensemble consisting of multiple multilayer perceptrons (MLPs), each MLP being responsible for one input feature thus making univariate decisions. 
The output values are summed and passed through a logistic sigmoid function for binary classification, or just summed for regression.
\cite{radenovic2022neural} propose a new sub-category of NAMs that uses a basis decomposition of shape functions. 
The basis functions are shared among all features, and are learned together for a specific task, making the model more scalable for large datasets with high-dimensional, sparse features.
 
\textbf{Differentiable Explicit Tree Methods:}
Although MLPs with piece-wise linear activation functions are already differentiable implicit trees \cite{aytekin2022neural}, there have been efforts to create other special architectures that results into explicit trees as follows.
\cite{joseph2022gate} proposes a gating mechanism for feature representation with attention based feature selection and builds on differentiable non-linear decision trees. 
In \cite{popov2019neural}, the authors present NODE, a deep learning architecture for tabular data that combines the strengths of ensembles of oblivious decision trees and end-to-end gradient-based optimization with multi-layer representation learning.

We find that univariate decisions and differentiable tree methods are promising in terms of interpretability and in line with our works.
Yet, the extracted decision trees in the literature are either soft\cite{popov2019neural}, or non-linear\cite{joseph2022gate} and univariate decisions made in \cite{agarwal2021neural}\cite{radenovic2022neural} are not still explainable.
On the contrary, LEURN provides an explainable and exact univariate decision tree.

Out of the above reviewed methods, the approach in \cite{popov2019neural} (NODE) is the most similar to ours.
Hence we find it important here to highlight some key differences.
In NODE, each layer contains multiple trees where each tree is trained via differentiable feature selection and decision making via entmax.
Note that made decisions and feature selections are soft and there is no exact corresponding tree. 
In our proposed method, we make hard decisions via quantized $tanh$ which results into exact univariate decision trees.
Another difference is that our feature selection is not explicit, but implicit through learnable feature importances.
In NODE,main computational power is spent on feature selection and thresholds for selected features are directly learned.
Instead, in our work, we spend main computational power on learning thresholds/rules as weighted combinations of a special embedding of previous responses and previous thresholds.
Most importantly, our method is able to explain exactly why a threshold was selected for a particular layer via linear contribution of previous thresholds and their responses.
Moreover our method has additional properties such as providing semantic similarity, generative usage, confidence scoring, etc.

\subsection{Post-processing Explainers}
The linear contributions of previous rules to others and final decision in LEURN is similar to the additive feature attributions mentioned in \cite{lundberg2017unified}. 
LIME \cite{ribeiro2016should} and SHAP \cite{lundberg2017unified} stand out as popular approaches as they approximate additive feature importances per sample.
These approaches are not novel neural network architectures, but they extract approximate explanations from existing neural networks. 
Thus their additive feature importances are local approximations.
LEURN differs from these approaches , because it is a novel neural network architecture and not a post-processing approximate explainer.
Moreover, LEURN's additive contributions are exact, not approximations, as they are a built-in feature of the architecture.
Finally, LIME and SHAP's additive contributions are only evident for features at decision level, whereas LEURN's additive contributions applies on every processing stage to intermediate rules on finding the next rule, as well as for the final decision.

\section{Proposed Method}
\label{sec:METHOD}

\subsection{Decision Tree Analysis of Vanilla Neural Networks}

A vanilla neural network with $n$ layers can be formulated in Eq. \ref{eq: ffd}.

\begin{equation}
\label{eq: ffd}
\begin{split}
NN(\textbf{x})=\bm{W}_{n-1}^T\sigma(\bm{W}_{n-2}^T\sigma(...\bm{W}_1^T\sigma(\bm{W}_0^T\bm{x}+\bm{\beta}_{0})
\\
+\bm{\beta}_{1} ... )+ \bm{\beta}_{n-2})+\bm{\beta}_{n-1}
\end{split}
\end{equation}

In Eq. \ref{eq: ffd}, $\bm{W}_i$ and $\bm{\beta}_i$ are respectively the weight matrix and bias vector of a network's $i^{th}$ layer, $\sigma$ is an activation function, and $\textbf{x}$ the input. 
Let us consider that $\sigma$ is a piece-wise linear activation.
Then, a layer makes decisions based on whether linear combinations of input features are larger than a set thresholds.
This fact was used to extract decision trees from neural networks in \cite{aytekin2022neural}, \cite{zhang2018tropical}, \cite{balestriero2017neural}, \cite{nguyen2020towards}, \cite{sudjianto2020unwrapping}.
The extracted decision trees differ from commonly used ones in two aspects, which we review as follows.

\subsubsection*{Exponentially Growing Branches}
Neural networks extract exponentially growing width trees due to shared rules embedded as layer weights.
In each layer $i$, the input's response to a rule is checked per filter $j$ and the effective operation depends which region of activation the response falls into. 
This results into $k^{\sum_{i}{m_i}}$ possible processing paths, where $k$ is the total number of linear regions in the activation and $m_i$ is the filter number in layer $i$.
This massive tree size hurts explainability in the global sense as it prevents to see all the decision mechanism at once. 
With today's very deep architectures it even becomes infeasible to store neural network extracted trees.
But, this feature doesn't prevent local explainability where the goal is to understand the decision mechanism of the neural network per sample.
At this point, we would also like to mention that exponential partitioning via shared rules has close connections to extrapolation, thus generalization \cite{aytekin2022neural}, hence this may be a favorable property to keep.

\subsubsection*{Mutivariate Decisions}

The decision rules of neural networks are multivariate.
This is obvious as the checked rule per filter is whether a linear combination of all features is larger than a value indicated by negative bias for that filter.
The set of multivariate rules extracted by vanilla neural networks are difficult to make sense of, as they mix features.
This is especially true for large number of features which is usually the case for modern neural networks.
Moreover, multivariate decisions make identification of redundant rules very difficult.
As stated in \cite{aytekin2022neural}, neural network extracted decision trees may consist of redundant rules, and the real depth/width may actually be a lot smaller than exponential formulation provided in previous subsection.
However, checking whether a set of multivariate rules encapsulates another set is actually very difficult compared to univariate rules.
Thus, in this work we wish to avoid this property of neural networks.

\begin{figure*}%
    \centering
    {{\includegraphics[width=14cm]{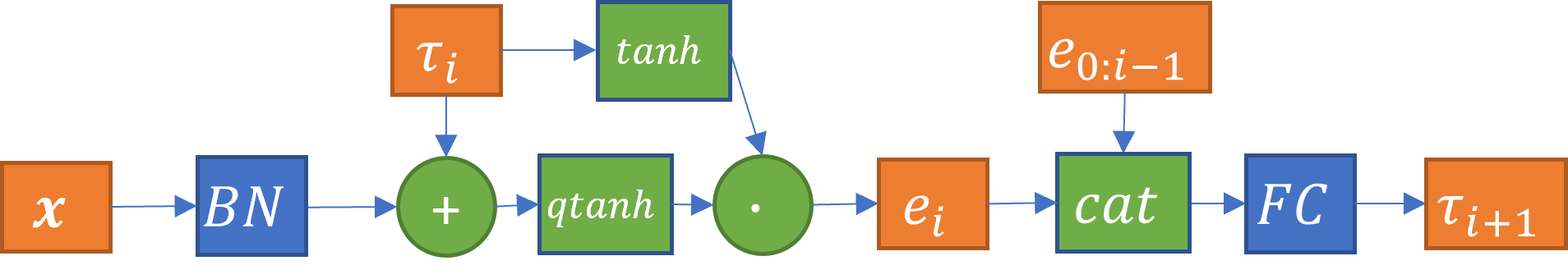} }}%
    \caption{LEURN's Rule Learning}
\label{fig:LEURN}
\end{figure*}

\subsection{LEURN: Learning Univariate Rules by Neural Networks}

In this section, we propose a special neural network architecture that results into univariate rules, while keeping generalization abilities of neural networks.
We will refer to this architecture as LEURN.
The main idea in LEURN is to learn univariate rules in the form of thresholds in each layer to be directly applied to the batch-normalized ($BN$) neural network input $\bm{x}$.
In layer $0$, we directly learn a rule vector $\bm{\tau}_{0}$ elements of which are separate rules for each input variable.
To find the rule vector for next layers, we employ the following.
For a layer $i$, first we find an indicator vector $\bm{s}_{i}$ which indicates the category of each input feature $j$ with respect to the rule $\tau_{ij}$.
This is achieved via a quantized $tanh$ activation which extracts thresholds around $\bm{\tau}_{i}$, and outputs unique values for each category whose boundaries are indicated by these thresholds. 
Then, we find an embedding $\bm{e}_{i}$ by element-wise multiplication of the indicator vector $\bm{s}_{i}$ with the activated threshold vector $tanh(\bm{\tau}_{i})$.  
This jointly encodes the thresholds used in the layer and how the input responded to them.
The next thresholds $\bm{\tau}_{i+1}$ are learned by a linear layer (FC) applied on the concatenated embeddings from previous layers $\bm{e}_{0:i}$ .
LEURN's rule learning is formulated in Eq. \ref{eq: LEURN} and visualized in Fig. \ref{fig:LEURN}.

\begin{equation}
\label{eq: LEURN}
\begin{split}
\bm{s}_{i} = qtanh(BN(\bm{x}) + \bm{\tau}_{i})
\\
\bm{e}_{i}=\bm{s}_{i}tanh(\bm{\tau}_{i})
\\
\bm{\tau}_{i+1} = \bm{W}_{i}^T\bm{e}_{0:i} + \bm{\beta}_{i}
\end{split}
\end{equation}

The output of the neural network $\bm{y}$ is calculated from all embeddings as follows.

\begin{equation}
\label{eq: last_layer}
\bm{y} = \gamma(\bm{W}_{n-1}^T\bm{e}_{0:n-1} + \bm{\beta}_{n-1})
\end{equation}

In Eq. \ref{eq: last_layer}, $\gamma$ is the final activation of the neural network.
This can be sigmoid or softmax for binary or multi-label classification, or simply identity for regression problems.

In summary, LEURN makes univariate decisions via $\tau$ rules and quantized $tanh$ with number of branches equal to $tanh$ quantization regions.
For every branch, a new rule or final decision is found via weighted linear combination of embeddings, so contribution of each rule and response is additive.



\subsubsection{Properties}
Next, we make a few observations about LEURN.

\textbf{Equivalent Univariate Decision Trees}: LEURN results into univariate decision trees.
For an input that is $n$-dimensional, in each layer there are $k^n$ possible indicator vectors which corresponds to $k^n$ branches per layer, where $k$ is the number of regions in quantized $tanh$.
Each branch is separated into another $k^n$ branches in the next layer.
Many of these branches are very unlikely to be utilized during training due to the limited number of data, so this rule sharing property helps generalization in inference time as the rules in unseen categories are made up from the seen ones.
Note that this final property is a general property of vanilla neural networks as well.

\textbf{Explainability}: LEURN is more explainable compared to vanilla neural networks.
In vanilla neural networks, the rules in the equivalent decision tree are in the form of multivariate inequalities, whereas in LEURN they are in the form of univariate inequalities.
Univariate inequalities are easier to make sense for humans.
LEURN is in this sense a white box which makes explanations in the following hypotethical example: Model checked and found that price/earning ratio (PER) of a company is smaller than 20 and operating income increase (OII) was more than 5$\%$. Based on this outcome, model then checked and found that PER of the company is larger than 15 and OII was less than 10$\%$. Therefore, model decided to invest 1k$\$$ on the company. Contribution of each rule-check to final invested money was as follows: PE ratio being smaller than 20: +1000$\$$ , operating income increase being more than 5$\%$: +1000$\$$, PE ratio being larger than 15: -500$\$$ , operating income increase being less than 10$\%$: -500$\$$. As a higher granularity of explainability, LEURN also provides how the  PE$<$20 and OII$>$5 rules linearly contributed to finding PE$>$15 and OII$<$10 rules. 

\textbf{Easy Architecture Search}: The last linear layer has fixed output units defined by the problem, the rest of the linear layers also have fixed output units which is equal to neural network input dimensions.
This makes neural architecture search easier as there are no hyperparameters in the form of number of features in a layer.
The architecture related hyperparameters are only the depth of the network and the number of quantization regions for $tanh$.

\textbf{Feature Selection}: LEURN can do feature selection as follows. 
Each embedding element is directly related to a particular input feature.
The embedded information is critical in finding next rules or in the output of the neural network.
Hence, if absolute value of the unit in $\bm{W}_{i}$ that corresponds to a particular feature is or close to zero, it is a clear indicator that the particular feature is uninformative, hence not selected in that layer.

\textbf{Feature Importance}: LEURN can extract global feature importance. 
Last layer's input is all the embeddings used throughout the neural network.
Feature importance can then be measured simply by checking the weighted (via $\bm{W}_{n-1}$) contribution of each embedding element -hence related feature and rule- in the classification, averaged over the training set.

\textbf{Pairwise Semantic Similarity}: LEURN can measure semantic similarity between two samples via using any popular distance metric on the embeddings of these samples.

\begin{figure*}
\begin{subfigure}{.24\textwidth}
  \centering
  \includegraphics[width=1\linewidth]{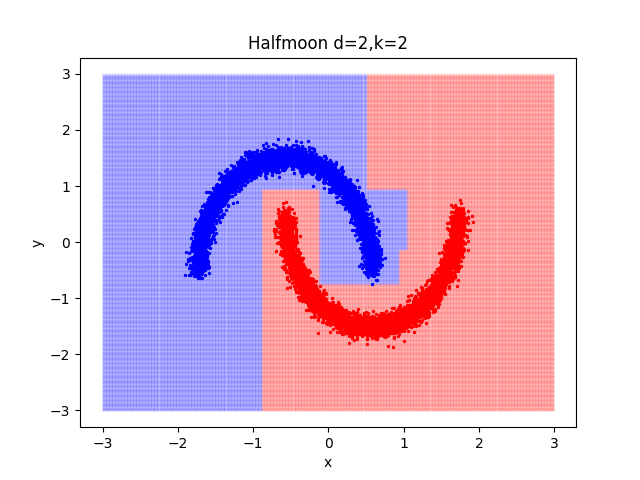}
  \caption{$k=2$}
  \label{fig:HMk2}
\end{subfigure}%
\begin{subfigure}{.24\textwidth}
  \centering
  \includegraphics[width=1\linewidth]{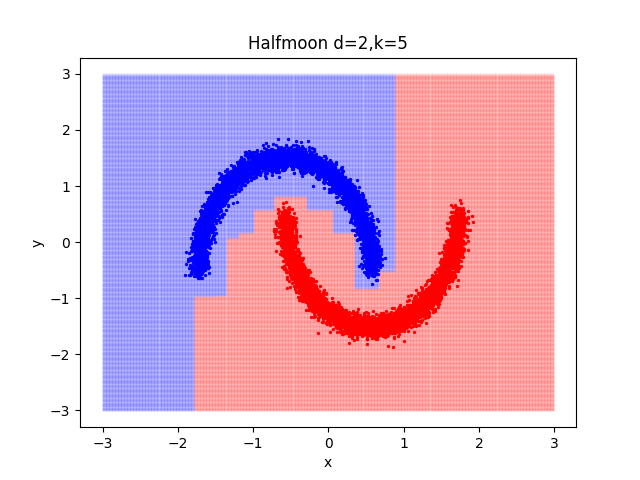}
  \caption{$k=5$}
  \label{fig:HMk5}
\end{subfigure}
\begin{subfigure}{.24\textwidth}
  \centering
  \includegraphics[width=1\linewidth]{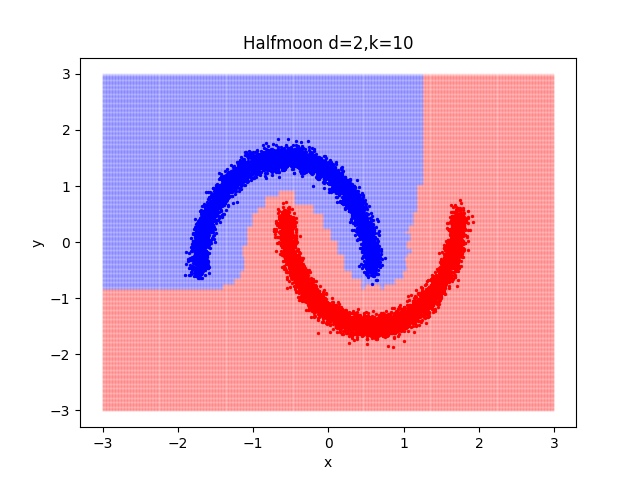}
  \caption{$k=10$}
  \label{fig:HMk10}
\end{subfigure}
\begin{subfigure}{.24\textwidth}
  \centering
  \includegraphics[width=1\linewidth]{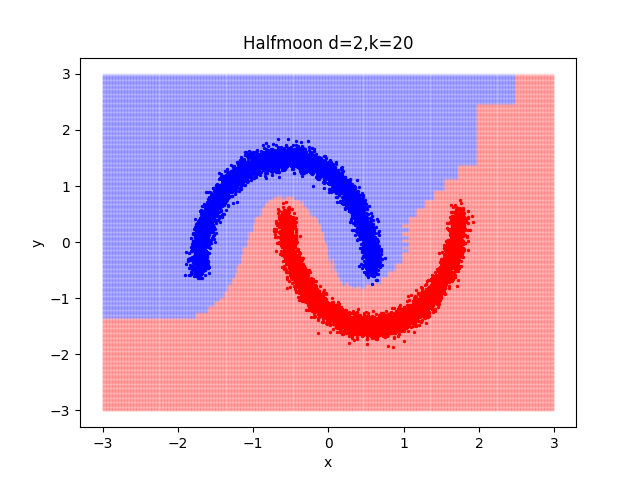}
  \caption{$k=20$}
  \label{fig:HMk20}
\end{subfigure}
\caption{Decision Boundaries of learned LEURN models with different $qtanh$ quantization regions $k$, for Half Moon dataset.}
\label{fig:HMk}
\end{figure*}

\textbf{Generative Usage}: LEURN can be directly used in a generative manner.
A training sample can be fed to the neural network and its univariate rules can be extracted.
This simply defines a category that belongs to that sample in the form of upper and lower limits for each feature in the input.
Then, one can generate a different sample from the same category by sampling randomly from the univariate rules per feature.
This is very difficult to do for vanilla neural networks, as it is harder to sample from multivariate inequalities.

\textbf{Parametrized Decision Smoothness}: Finally LEURN provides a controlled smoothness in model predictions based on the number of quantized regions $k$ in the $qtanh$ function. 
As $k$ grows, decision boundaries become smoother, this hyperparameter is useful to provide alternatives to different datasets based on their properties.

\section{Experimental Results}
\subsection{Preliminary Experiments on Toy Data}
\subsubsection{Parametrized Decision Smoothness}

In \cite{grinsztajn2022tree}, authors stated three key differences between neural network based methods and tree-based methods. 
These were: decision boundary smoothness, robustness to uninformative features and rotation invariance.
In this section, we experiment on the behaviour of LEURN on a toy dataset and observe that with different $qtanh$ quantization regions $k$, LEURN behaves differently in terms of the above three aspects.

For all experiments in this section, we use Half Moon dataset with 10000 training samples and 1000 validation samples. 
Accuracies are reported on best validation performance in a training.

First, we investigate decision boundary smoothness. 
For all LEURNs, we have used fixed depth of $d=2$ and experimented on $qtanh$ quantization regions in following set: $k\in\{2,5,10,20\}$.
As it can be observed from Fig. \ref{fig:HMk}, as the quantization regions grow, decision boundary becomes smoother.
Note that this result is trivially evident without this experiment, but we still provide these figures for completeness.

Second, we rotate the Half Moon dataset with angles in the following set: $\{0,11.25,22.5,45\}$.
We average results of 10 trainings for each quantization region in: $\{2,5,10,20\}$.
To provide a challenging case, we have fixed network depth to 1 in this experiment.
As it can be seen from Table \ref{tab:rotexp} , as the quantization regions grow, LEURN becomes more robust to rotation, which is a vanilla neural networks-like behaviour according to \cite{grinsztajn2022tree}.
Smaller quantization regions struggle to be robust to rotation, similar to decision tree behaviour, according to \cite{grinsztajn2022tree}.

\begin{table}
\tiny
\begin{adjustbox}{width=\linewidth}
\begin{tabular}{|c|c|c|c|c|}

 \hline
  k & 2 & 5 & 10 & 20 \\
 \hline
  $\mu$ & 96.79 & 98.44 & 98.76 & 99.01  \\
 \hline
  $\sigma$ &  3.22 & 1.91 & 1.64 & 1.26 \\
 \hline
\end{tabular}
\end{adjustbox}
\caption{\label{tab:rotexp} Mean and standard deviances of LEURNs with different $qtanh$ quantization regions, across different rotations.}
\end{table}

\begin{figure*}
\begin{subfigure}{.3\textwidth}
  \centering
  \includegraphics[width=1\linewidth]{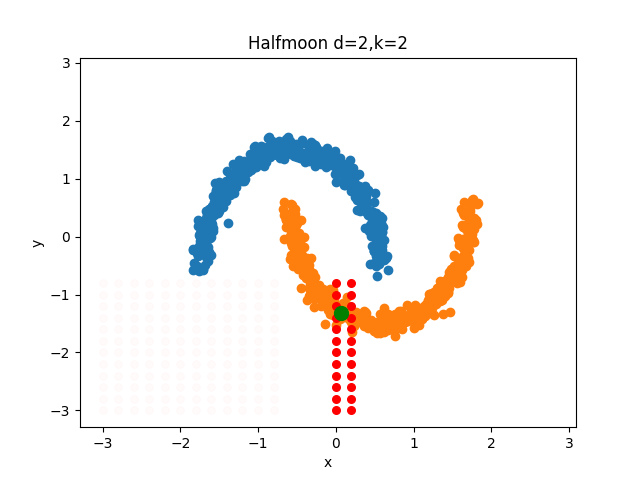}
  \caption{$d=2, k=2$}
  \label{fig:HMk2}
\end{subfigure}%
\hfill
\begin{subfigure}{.3\textwidth}
  \centering
  \includegraphics[width=1\linewidth]{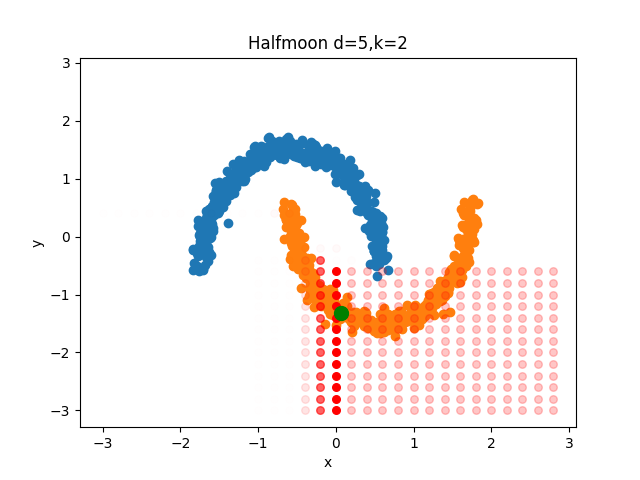}
  \caption{$d=5, k=2$}
  \label{fig:HMk5}
\end{subfigure}
\hfill
\begin{subfigure}{.3\textwidth}
  \centering
  \includegraphics[width=1\linewidth]{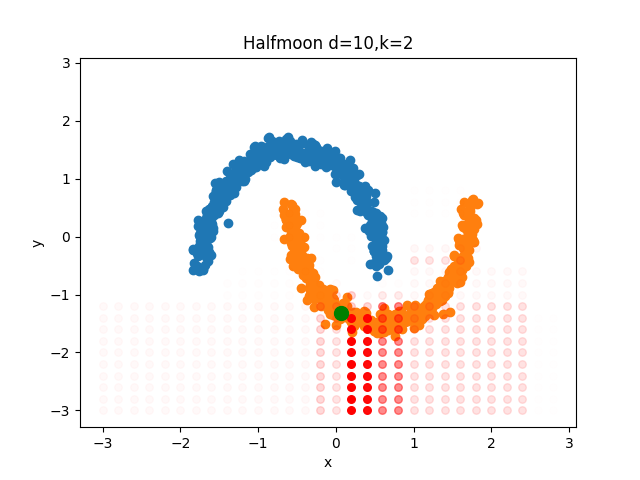}
  \caption{$d=10, k=2$}
  \label{fig:HMk10}
\end{subfigure}
\\
\begin{subfigure}{.3\textwidth}
  \centering
  \includegraphics[width=1\linewidth]{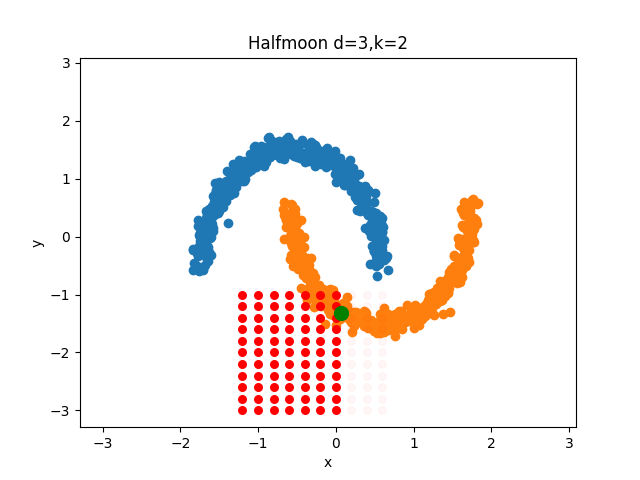}
  \caption{$d=3, k=2$}
  \label{fig:HMk20}
\end{subfigure}
\hfill
\begin{subfigure}{.3\textwidth}
  \centering
  \includegraphics[width=1\linewidth]{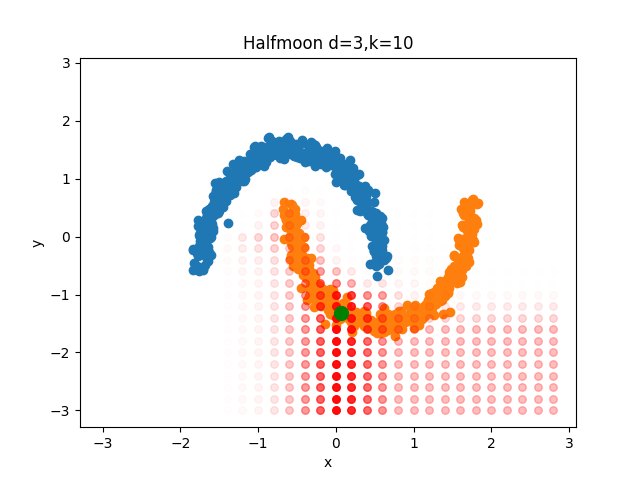}
  \caption{$d=3, k=10$}
  \label{fig:HMk2}
\end{subfigure}%
\hfill
\begin{subfigure}{.3\textwidth}
  \centering
  \includegraphics[width=1\linewidth]{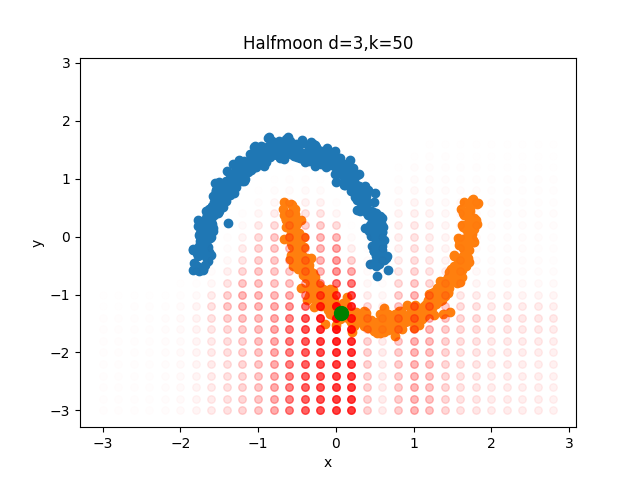}
  \caption{$d=3, k=50$}
  \label{fig:HMk5}
\end{subfigure}

\caption{Embedding similarities to reference sample for LEURN models with different $qtanh$ quantization regions $k$ and depths $d$, for Half Moon dataset.}
\label{fig:SemSimil}
\end{figure*}

Finally, we experiment on uninformative features. 
We add 10 additional randomly generated features to Half Moon dataset in order to provide a challenging case with dominating uninformative features. 
We have used a fixed depth of 3 in this experiment so that all variants are able to provide perfect score without uninformative features.
We have observed that all LEURN variants with different quantization regions can successfully handle uninformative features by providing perfect score on validation set.

We conclude from these three experiments that, LEURN provides smoother boundaries (MLP-like) with rotation invariant performance for higher quantization regions, and sharper boundaries with rotation variant performance for lower quantization regions (DT-like). 
For all cases, LEURN was found to be robust to uninformative features.

%
%

\subsubsection{Global Feature Importance}

We repeat our last experiment in previous section with $k=2,d=2$.
We measure the feature importance as the contribution in the final classification.
The process is as follows.
The input of the last layer is a concatenation of previous embeddings used throughout LEURN.
Each embedding unit corresponds to an information from a particular feature.
Thus the value of the multiplied embedding unit with corresponding weight unit in the last layer is treated as that feature's importance.
As there are multiple embeddings per feature, we first sum each contributions, then take absolute value of the sum to get the final importance value.
Note that these values are calculated over the training set.
With this method, we find that informative features in the last experiment, are given at least 4.34 times more importance than uninformative features.

Feature importance can also be calculated in all intermediate fully connected layers.
It can also be calculated locally, i.e. for a particular sample instead of over entire training set.

\subsubsection{Semantic Similarity, Generative Usage and Confidence Scores}

In this section we use the RBF kernel response of a pair of LEURN embeddings as a similarity score between corresponding pair of samples.
We use Half Moon dataset with several $k$ and $d$ hyperparameter choices.
In Fig. \ref{fig:SemSimil}, we select a reference sample from training dataset indicated by the green dot, and visualize semantically similar regions with red dots, we set opacity parameter to similarity value, hence stronger red dots means more similar samples to the reference.
Smaller $k$ and $d$ results into more distinct and local categories where everything in the category is strongly similar to the reference and similarity sharply stops in category boundaries.
On the contrary, larger $k$ and $d$ results into distributed and non-local similarities.
One can select the desired behaviour based on application.

\begin{figure}
  \begin{subfigure}{.23\textwidth}
  \centering
  \includegraphics[width=1\linewidth]{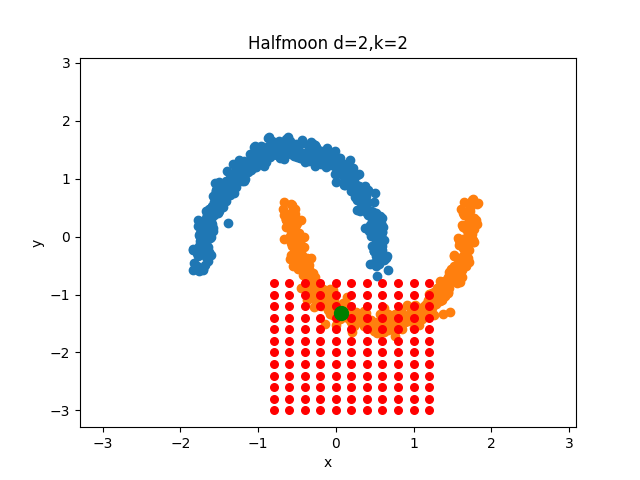}
  \caption{$d=2, k=2$}
  \end{subfigure}
  \begin{subfigure}{.23\textwidth}
  \centering
  \includegraphics[width=1\linewidth]{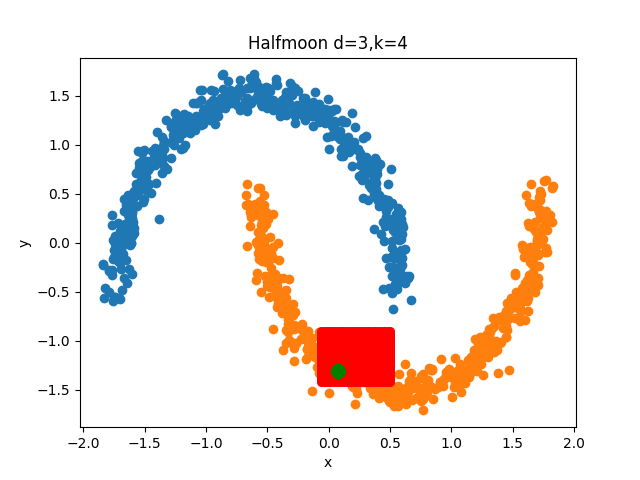}
  \caption{$d=3, k=4$}
  \end{subfigure}
\caption{Generated samples from region that reference sample (green) falls into}
\label{fig:Gen}
\end{figure}

As LEURN generates univariate trees, there is a distinct category for each sample, this category can be easily found via storing the thresholds that are employed throughout LEURN and finding closest upper and lower boundaries from these thresholds. 
Note that these thresholds are found through $\tau$ vectors and $qtanh$ quantization regions.
Every sample in these boundaries have exactly 0 embedding distance, hence 1 similarity to the reference sample. 
An example of generated samples can be ovserved in Fig \ref{fig:Gen}. 
We observe that $h$ and $d$ grows, the regions gets tighter.
This is expected due to the fact that both $k$ and $d$ partitions the space into more regions. 
At first, this can be confused to conflict results of Fig. \ref{fig:SemSimil}, however we remind the reader that most red dots in Fig. \ref{fig:SemSimil} do not have exactly 1 similarity, thus -although semantically similar- not exactly in same category with reference sample.
Moreover some regions may vary depending on the category that each training finds due to the random initialization.

\begin{figure}
  \begin{subfigure}{.23\textwidth}
  \centering
  \includegraphics[width=1\linewidth]{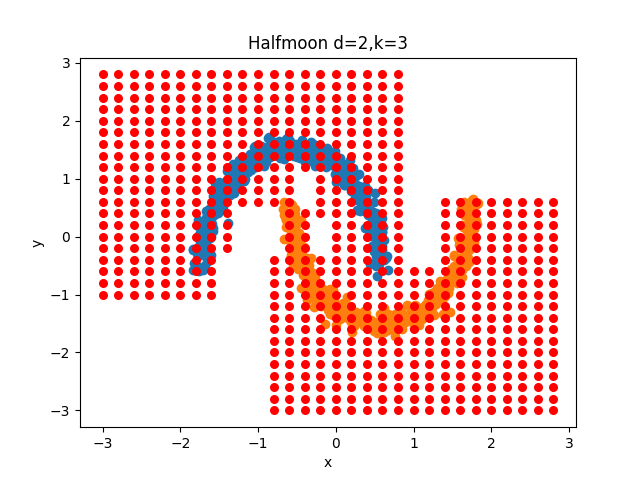}
  \caption{$d=2, k=3$}
  \end{subfigure}
  \begin{subfigure}{.23\textwidth}
  \centering
  \includegraphics[width=1\linewidth]{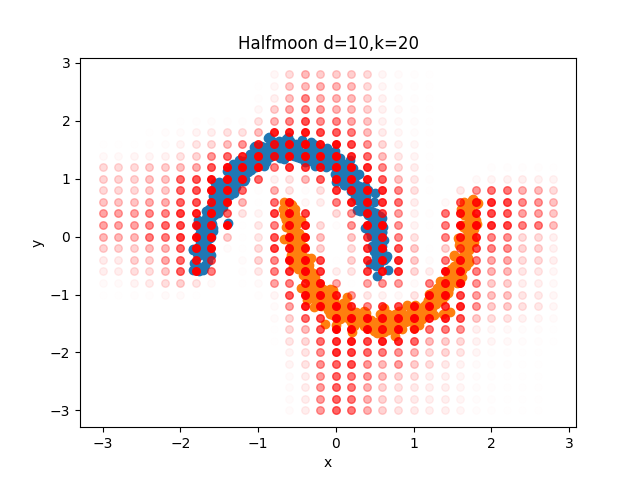}
  \caption{$d=10, k=20$}
  \end{subfigure}
\caption{Confidence scores of samples}
\label{fig:Conf}
\end{figure}

\begin{figure}%
    \centering
    {{\includegraphics[width=8.5cm]{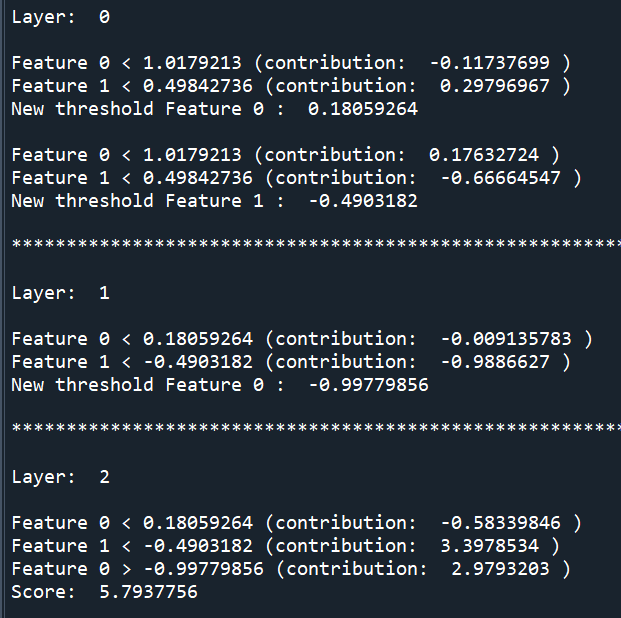} }}%
    \caption{Explanation of LEURN for Half-Moon Dataset}
\label{fig:Explain}
\end{figure}

\begin{figure*}
\begin{subfigure}{.33\textwidth}
  \centering
  \includegraphics[width=1\linewidth]{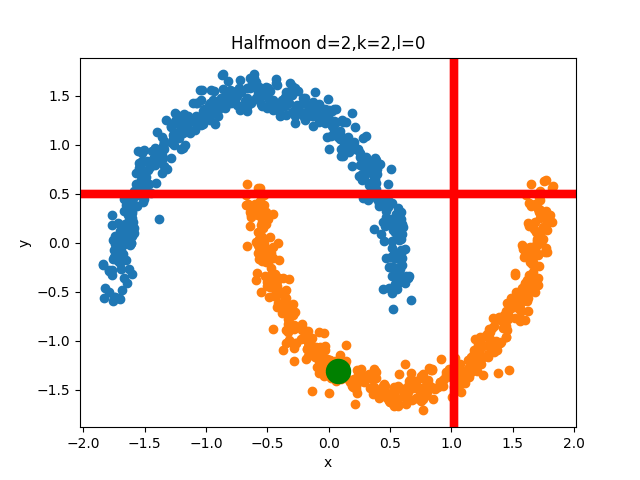}
  \caption{Layer 0}
  \label{fig:RuleL0}
\end{subfigure}%
\begin{subfigure}{.33\textwidth}
  \centering
  \includegraphics[width=1\linewidth]{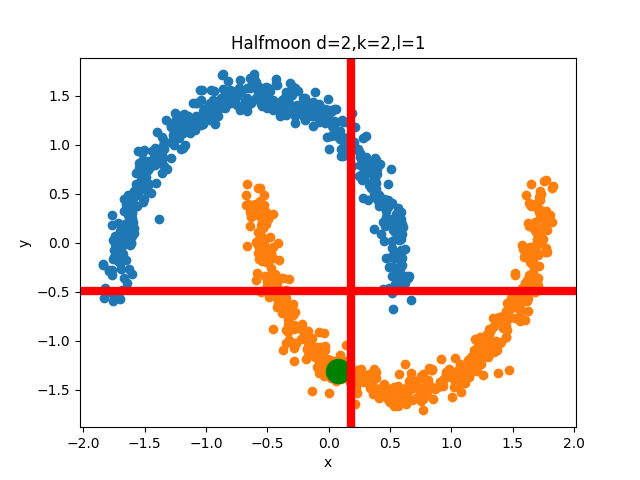}
  \caption{Layer 1}
  \label{fig:RuleL1}
\end{subfigure}
\begin{subfigure}{.33\textwidth}
  \centering
  \includegraphics[width=1\linewidth]{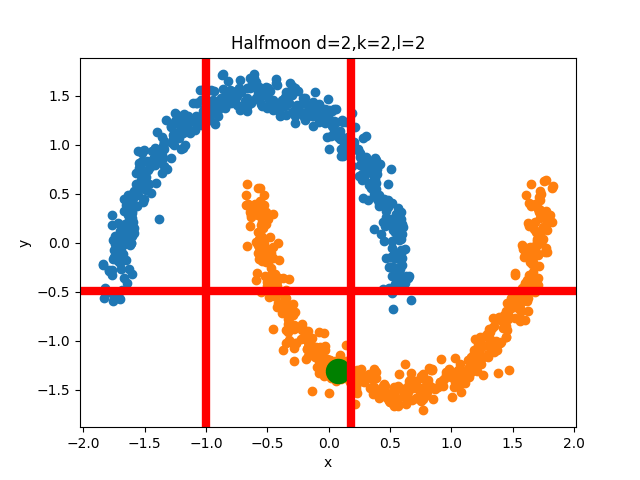}
  \caption{Layer 2}
  \label{fig:RuleL2}
\end{subfigure}
\caption{Utilized univariate rules in every layer.}
\label{fig:Rules}

\end{figure*}

The embedding similarity can also be used to assign confidence scores to predictions in test set.
We simply find the maximum similarity that a test sample has to training dataset in the nearest neighbour sense.
This value is directly used as a confidence score.
In Fig. \ref{fig:Conf}, we visualize confidence scores via setting confidence score to opacity.
Similar to our other observations in this section, lower $k$ and $d$ values result into larger regions that have high confidence whereas higher $k$ and $d$ values result into tighter categories, thus lower confidence scores in extrapolated regions.

\subsection{Explainability}

As observed from Eq. \ref{eq: LEURN} and \ref{eq: last_layer}, contribution of each embedding in finding the next threshold and the final decision is simply linear.
As an embedding is a scalar which encodes a previous threshold and input response to it, it is possible to write linear weights of each previous threshold in finding next threshold according to the input response.
There can be sometimes redundant thresholds found in intermediate layers.
For the sake of clarity, we handle these as follows.
We check whether a previous threshold is above/below the upper/lower limits that are found via previous thresholds.
If so, this means that threshold is redundant.
Then, the contribution of that threshold in the network is added to the contribution of the previous threshold that defines upper/lower limits.
This provides a much clearer and human readable explanation.
Note that it also keeps original processing with no alteration, it is just another way to reformulate the neural network/decision tree.
Finally, we by-pass the thresold finding mechanisms in the neural network that results into redundant thresholds to further simplify the tree, while keeping it equivalent to the original one.
Finally, we distribute the contribution of the biases to each rule equally.

Next, we test explainability module on two cases: Half-Moon and Adult Census Income (OpenML \cite{vanschoren2014openml} $\#1590$) datasets.

\begin{table*}
\tiny
\resizebox{\textwidth}{!}{
\begin{tabular}{|c|c|c|c|} 
 \hline
 Feature & LEURN Rule & LEURN Scores & SHAP Scores\\ 
 \hline\hline
 workclass & State-gov & 0.0131&	-0.2027\\ 
 \hline
 education &	Bachelors	&0.1897 	&-0.2514\\
 \hline
 marital-status&	Married-civ-spouse&	0.1719&	0.7643\\
 \hline
 occupation&	Exec-managerial&	0.2967&	0.1796\\
 \hline
 relationship & Husband &	0.0985 &	0.2535\\
 \hline
 race&	White	&0.0117&	-0.0982\\
 \hline
 sex&	Male&	0.0162	&0.1020\\
 \hline
 native-country&	USA&	0.1193	&-0.9945\\
 \hline
age	&$71.75>X>66.79$&	0.1234&	0.9786\\
\hline
 fnlwgt&	$14254151.00>X>-612719168.00$&	-0.0034&	-0.0244\\
 \hline
education num	&$13.13>X>12.83$	&0.2005&	0.5779\\
\hline
 capital-gain&	$7433545.31>X>-1413678.12$	&-0.0882&	-0.2000\\
 \hline
capital loss	&$238.01>X>-10163.00$&	-0.0609	&-0.0912\\
\hline
hours per week&$	40.52>X>39.91$&	-0.0342&	0.0619\\
\hline
\end{tabular}
}
\caption{LEURN Rules, LEURN Scores and SHAP Scores assigned to features of a sample from Adult Income Census Dataset}
\label{LEURN_ADULT}
\end{table*}

\subsubsection*{Toy Data}

Next, we explain the decisions of a LEURN trained for the Half Moon dataset with $k=2$ and $d=2$.
The output of the explanation described in the above paragraph is visualized in Fig. \ref{fig:Explain} which is the direct output of the explainer module.
The corresponding univariate rules utilized in each layer of LEURN are visualized in Fig. \ref{fig:Rules}.

As it can be observed from Fig. \ref{fig:Explain} and Fig. \ref{fig:Rules}, in the first layer, network checks rules for input features separately.
Note that these rules $(x ,1.018),(y,0.498)$ are same for any sample/category and these are the only directly learned thresholds in the network.
Based on the position of our sample (indicated by green dot in Fig. \ref{fig:Rules}) w.r.t to these rules, network decides to check other rules: $(x ,0.181),(y,-0.490)$ respectively. 
This decision is sensible, since from the position of our sample, it is not clear to which class it belongs to with previous thresholds.
The new thresholds tighten the boundary, but still are not sufficient due to a few still-included blue samples in the bounded region.
So finally, $(x ,-0.998)$ rule is found which is enough to accurately classify our sample.

We also check the contributions of each sensible threshold in the final classification score (last section of Fig. \ref{fig:Explain}).
Note that positive score means orange class here.
$x<0.181$ results into a negative contribution in class orange.
This is meaningful, because most samples at this region belongs to blue class.
$y<-0.490$ results into a positive contribution in class orange.
This is meaningful, because most samples at this region belongs to orange class.
$x>-0.998$ results into a positive contribution in class orange.
This is meaningful, because all samples at this region belongs to orange class.

It is also interesting to examine how the previous thresholds contributed to finding the new thresholds. 
Let us examine the case in Layer 1.
$x<0.181$ pushes to check $(x,-0.009)$ rule whereas $y<-0.490$ pushes to check $(x,-0.989)$ rule.
Note that both are separately enough to classify our sample.

A limitation of the method in terms of explainability is that the rules applied on the input features and their linear contribution to either finding the next rule or making the final decision still needs human effort for interpretation/speculation.
For example, there is no explicit explanation given by the network for why $y<-0.490$ results into a positive contribution in class orange, but we interpret/speculate it.
Still, LEURN simplifies the interpretability problem to an unprecedentedly easy level for humans.
We believe best use of LEURN is to assist a human expert in the application field where the human expert is required to verify and make sense of the explanations that LEURN makes in the form of Fig. \ref{fig:Explain}.

\begin{table*}
\resizebox{\textwidth}{!}{
\begin{tabular}{||c | c c c c c c c c c c|c|} 
 \hline
 Method / OpenML ID & 31 & 44 & 1017 & 1111 & 1487 & 1494 & 1590 & 4134 & 42178 & 42733 & Average \\ 
 \hline\hline
 RandomForest & 0.778&  0.986 & 0.798 & 0.774 & 0.910 & 0.928 & 0.908 & 0.868 & 0.840 & 0.670 & 0.846\\ 
 \hline
 ExtraTrees & 0.764 & 0.986 & 0.811 & 0.748 & 0.921 & 0.935 & 0.903 & 0.856 & 0.831 & 0.659 & 0.841 \\
 \hline
 KNeighborsDist & 0.501 & 0.873 & 0.722 & 0.517 & 0.741 & 0.868 & 0.684 & 0.808 & 0.755 & 0.576 & 0.705 \\
 \hline
 KNeighborsUnif & 0.489 & 0.847 & 0.712 & 0.516 & 0.734 & 0.865 & 0.669 & 0.790 & 0.764 &  0.578 &  0.696\\
 \hline
 LightGBM& 0.751& 0.989 &0.807 &0.803 &0.911& 0.923& 0.930& 0.860& 0.853 &0.683& 0.851 \\ 
 \hline
 XGBoost& 0.761 &0.989& 0.781 &0.802& 0.903& 0.915& 0.931 &0.864& 0.854& 0.681 &0.848 \\ 
 \hline
 CatBoost& 0.788 &0.987& 0.838& 0.818& 0.914 &0.931& 0.930 &0.858& 0.856& 0.686 &0.860 \\ 
 \hline
 MLP& 0.705& 0.980& 0.745& 0.709 &0.913 &0.932& 0.910& 0.818 &0.841 &0.647& 0.820 \\ 
 \hline
 TabNet& 0.472 &0.978 &0.422& 0.718&0.625& 0.677& 0.917 &0.701& 0.830 &0.603& 0.694 \\ 
 \hline
 TabTransformer& 0.764 &0.980 &0.729 &0.763 &0.884 &0.913& 0.907& 0.809 & 0.841& 0.638& 0.823 \\ 
 \hline
 SAINT& 0.790& 0.991 &0.843 &0.808 &0.919& 0.937& 0.921 &0.853& 0.857 &0.676 &0.859 \\ 
 \hline
 LEURN& 0.772& 0.985 &0.817 & 0.810 &0.915& 0.930& 0.912 &0.858& 0.848 &0.649 &0.850 \\ 
 \hline

\end{tabular}
}
\caption{Comparison to state-of-the-art in Binary Classification Datasets}
\label{SOTA_Binary}
\end{table*}

\begin{table*}
\resizebox{\textwidth}{!}{
\begin{tabular}{||c | c c c c c c c c c c|c|} 
 \hline
 Method / OpenML ID & 188 &1596 &4541 &40664 &40685& 40687 &40975 &41166 &41169 &42734 & Average \\ 
 \hline\hline
 RandomForest&  0.653 & 0.953 & 0.607&  0.951&  0.999&  0.697&  0.967 & 0.671&  0.358 & 0.743&  0.760\\ 
 \hline
 ExtraTrees &0.653 &0.946 &0.595 &0.951 &0.999 &0.697 &0.956& 0.648& 0.341& 0.736 &0.752\\
 \hline
 KNeighborsDist &0.442& 0.965& 0.491& 0.925& 0.997 &0.720 &0.893 &0.620 &0.205& 0.685& 0.694 \\
 \hline
 KNeighborsUnif & 0.422 & 0.963 & 0.489 & 0.910 & 0.997 & 0.739 & 0.887&  0.605 & 0.189&  0.693 & 0.689\\
 \hline
 LightGBM&0.667 &0.969& 0.611& 0.984& 0.999 &0.716& 0.981& 0.721 &0.356& 0.754 &0.776 \\ 
 \hline
 XGBoost& 0.612 &0.928& 0.611 &0.984& 0.999 &0.730 &0.984 &0.707 &0.356& 0.752& 0.766\\ 
 \hline
 CatBoost& 0.667& 0.871 &0.604& 0.986& 0.999& 0.730 &0.962& 0.692 &0.376& 0.747 &0.763\\ 
 \hline
 MLP&0.388 &0.915& 0.597& 0.992 &0.997 &0.682& 0.984& 0.707& 0.378& 0.733 &0.737 \\ 
 \hline
 TabNet& 0.259 &0.744& 0.517& 0.665& 0.997& 0.275& 0.871& 0.599& 0.243 &0.630 &0.580 \\ 
 \hline
 TabTransformer& 0.660 &0.715& 0.601& 0.947& 0.999 &0.697& 0.965 &0.531& 0.352 &0.744& 0.721 \\ 
 \hline
 SAINT& 0.680 &0.946& 0.606 &1.000 &0.999& 0.735 &0.997 &0.701& 0.377 &0.752& 0.779\\ 
 \hline
 LEURN& 0.644& 0.963 & 0.595 & 0.995 &0.997& 0.768& 0.994 &0.654& 0.343 &0.746 & 0.769\\ 
 \hline

\end{tabular}
}
\caption{Comparison to state-of-the-art in Multi-Class Classification Datasets}
\label{SOTA_Multi}
\end{table*}

\begin{table*}
\resizebox{\textwidth}{!}{
\begin{tabular}{||c | c c c c c c c c c c|c|} 
 \hline
 Method / OpenML ID & 1422  &541 & 42563  &42571  &42705  &42724 & 42726  &42727  &42728 & 42729  \\ 
 \hline\hline
 RandomForest&  0.027& 17.814& 37085.577& 1999.442& 16.729 &12375.312 &2.476& 0.149& 13.700& 1.767\\ 
 \hline
 ExtraTrees &0.027& 19.269& 35049.267& 1961.928& 15.349& 12505.090& 2.522& 0.147 &13.578& 1.849\\
 \hline
 KNeighborsDist &0.029& 25.054& 46331.144& 2617.202 &14.496& 13046.090& 2.501& 0.167& 13.692& 2.100 \\
 \hline
 KNeighborsUnif & 0.029 &24.698 &47201.343& 2629.277& 18.397 &12857.449& 2.592& 0.169 &13.703 &2.109\\
 \hline
 LightGBM&0.027 &19.871 &32870.697 &1898.032& 13.018& 11639.594& 2.451& 0.144& 13.468 &1.958 \\ 
 \hline
 XGBoost& 0.028 &13.791& 36375.583 &1903.027 &12.311 &11931.233& 2.452& 0.145 &13.480&1.849\\ 
 \hline
 CatBoost& 0.027 &14.060 &35187.381 &1886.593& 11.890& 11614.567& 2.405& 0.142& 13.441& 1.883\\ 
 \hline
 MLP&0.028& 22.756 &42751.432 &1991.774& 15.892& 11618.684& 2.500& 0.162 &13.781& 3.351 \\ 
 \hline
 TabNet& 0.028 &22.731& 200802.769 &1943.091 &11.084 &11613.275& 2.175& 0.183 &16.665 &2.310 \\ 
 \hline
 TabTransformer& 0.028 &21.600 &37057.686& 1980.696& 15.693& 11618.356& 2.494 &0.162& 12.982& 3.259 \\ 
 \hline
 SAINT& 0.027& 11.661 &33112.387 &1953.391 &10.282 &11577.678& 2.113& 0.145& 12.578 &1.882\\ 
 \hline
 LEURN& 0.028&14.052 & 30407.997 &2061.0534& 10.666&11900.791 &2.194& 0.155 &13.345 &1.936  \\ 
 \hline

\end{tabular}
}
\caption{Comparison to state-of-the-art in Regression Datasets}
\label{SOTA_Reg}
\end{table*}

\subsubsection*{Adult Census Income Dataset}

In this section we train LEURN on Adult Income Census dataset. 
Adult Income Census dataset contains samples having features of an adult's background, and the task is to predict whether the adult earns more than 50000$\$$.

In Table \ref{LEURN_ADULT}, we provide the rules and explanations LEURN finds for a sample. 
We also compare the additive feature contributions of LEURN and SHAP. 
Note that SHAP is applied to trained LEURN model. 

First we analyze individual feature contributions. 
The sample is classified as earning more than 50000$\$$ and the largest contributing positive factor was executive managerial job title which makes excellent sense.
Other contributing factors were: Education number, Bachelor degree education and married civilian marital status which are sensible.
Note that first two are repeating features in the dataset, and both are given high importance which is consistent.
An interesting observation is that LEURN finds redundant rules for final weight (fnlwgt) and capital gain features as the found rules that sample belong to exceeds the minimum and maximum values for these features in the dataset.
The contributions coming from these redundant rules may be considered as a bias of the particular category.
Finally, we observe a non-sensible contribution coming from capital loss. 
Although the capital loss interval here points towards no loss case, the model assigns a negative contribution here which maybe does not make sense.
It is important to note here that, it is a very important feature to present these rules to humans as non-sensible decisions such as this particular one can be directly visible during monitoring.
Although in this particular case this does not alter the final decision, it may be important in some other scenarios.

As the LEURN scores are directly drawn from each feature's and rule's additive contribution in the final layer, they are exact.
When this is considered, one would expect that SHAP would extract similar importances per feature.
However, it is clearly evident from Table \ref{LEURN_ADULT} that SHAP explanations deviate considerably from these exact contributions.

Some key differences we see from Table \ref{LEURN_ADULT} is that SHAP finds native country being USA is one of the biggest negative factors, whereas in reality contribution from that feature is positive. 
SHAP also shows that a Bachelor degree, a state government job and being white are negative indicators, whereas in LEURN's exact contributions they are positive. Another interesting observation is SHAP gives more importance to being married to a civilian spouse than executive managerial job title, which does not make sense.
We observe similar behaviours across the dataset.

\subsection{Comparison to State-of-the-Art in Tabular Data}

In this section, we provide comparisons to popular methods for tabular data.
The methods, datasets and evaluation procedure follows \cite{somepalli2021saint} as we use their code for retrieving the datasets, preprocessing and splits.
Therefore, the performance of competitor methods have been copy-pasted from  \cite{somepalli2021saint}.
Compared methods are Random Forests \cite{breiman2001random}, Extra Trees \cite{geurts2006extremely}, k-NN \cite{altman1992introduction}, LightGBM \cite{ke2017lightgbm}, XGBoost \cite{chen2016xgboost}, CatBoost \cite{dorogush2018catboost}, TabNet \cite{arik2021tabnet}, TabTransformer \cite{huang2020tabtransformer} and SAINT \cite{somepalli2021saint}.
In Tables \ref{SOTA_Binary}, \ref{SOTA_Multi} and \ref{SOTA_Reg}, one can find comparisons of LEURN with state-of-the-art methods.
Following \cite{somepalli2021saint}, we have used datasets that are available in OpenML \cite{vanschoren2014openml}, and used OpenML identifiers in the comparison tables.
The scores are given in area under receiver operating characteristics curve, accuracy and root mean square error for binary classification, multiclass classification and regression problems respectively.

For all trainings, we perform automatic hyperparameter selection as follows.
LEURN has three hyperparameters: depth ($d$), tanh quantizated region number ($k$) and dropout rate ($r$).
We define search intervals $d \in \{0,1,2,5,10\}$, $k \in \{1,2,5,10\}$, $r \in \{0,0.1,0.3,0.5,0.7,0.9\}$.
During hyperparameter search, first we set $k=1, r=0.9$ (most regularized case) and find best depth.
We sweep $d$ from smallest to largest and stop search when performance metric becomes worse.
Next, with $d=d_{best}$ and $r=0.9$ set, we sweep $k$ from smallest to largest and stop when there is no improvement.
Finally, we set $d=d_{best}$ and $k=k_{best}$, and sweep $r$ from largest to smallest and stop when there is no improvement.
In the above process, the main idea is to start from most regularized case, and check performance improvement when regularization is softened in a controllable way. 
The search order of $d,k,r$ is emprical.
When performance metric is checked in each stage, we perform 5 trainings with random training and validation sets where best performance is found via best validation error. 

Once hyperparameter search is complete, we perform 20 trainings with selected hyperparameters on random training, validation and test splits.
We save best performing model on validation data, and report test performance averaged over 20 trainings.
The split ratios follow \cite{somepalli2021saint}.

As one can observe, LEURN is comparable and sometimes favorable to state-of-the-art methods, while having explainability advantages.

\section{Conclusion}

We have introduced LEURN: Learning explainable univariate rules with neural networks.
We have shown that LEURN makes human explainable decisions by its special design that results into learning rules with additive contributions.
Several other advantages of LEURN was highlighted and tested on a toy dataset.
LEURN was tested on 30 public tabular datasets, and it was found comparable to state-of-the-art methods.

{\small
\bibliographystyle{ieee_fullname}
\bibliography{egbib}

\begin{thebibliography}{10}\itemsep=-1pt

\bibitem{agarwal2021neural}
Rishabh Agarwal, Levi Melnick, Nicholas Frosst, Xuezhou Zhang, Ben Lengerich,
  Rich Caruana, and Geoffrey~E Hinton.
\newblock Neural additive models: Interpretable machine learning with neural
  nets.
\newblock {\em Advances in Neural Information Processing Systems},
  34:4699--4711, 2021.

\bibitem{ahmed2016network}
Karim Ahmed, Mohammad~Haris Baig, and Lorenzo Torresani.
\newblock Network of experts for large-scale image categorization.
\newblock In {\em European Conference on Computer Vision}, pages 516--532.
  Springer, 2016.

\bibitem{altman1992introduction}
Naomi~S Altman.
\newblock An introduction to kernel and nearest-neighbor nonparametric
  regression.
\newblock {\em The American Statistician}, 46(3):175--185, 1992.

\bibitem{arik2021tabnet}
Sercan~{\"O} Arik and Tomas Pfister.
\newblock Tabnet: Attentive interpretable tabular learning.
\newblock In {\em Proceedings of the AAAI Conference on Artificial
  Intelligence}, volume~35, pages 6679--6687, 2021.

\bibitem{aytekin2022neural}
Caglar Aytekin.
\newblock Neural networks are decision trees.
\newblock {\em arXiv preprint arXiv:2210.05189}, 2022.

\bibitem{bahri2021scarf}
Dara Bahri, Heinrich Jiang, Yi Tay, and Donald Metzler.
\newblock Scarf: Self-supervised contrastive learning using random feature
  corruption.
\newblock {\em arXiv preprint arXiv:2106.15147}, 2021.

\bibitem{balestriero2017neural}
Randall Balestriero.
\newblock Neural decision trees.
\newblock {\em arXiv preprint arXiv:1702.07360}, 2017.

\bibitem{borisov2022deep}
Vadim Borisov, Tobias Leemann, Kathrin Se{\ss}ler, Johannes Haug, Martin
  Pawelczyk, and Gjergji Kasneci.
\newblock Deep neural networks and tabular data: A survey.
\newblock {\em IEEE Transactions on Neural Networks and Learning Systems},
  2022.

\bibitem{breiman2001random}
Leo Breiman.
\newblock Random forests.
\newblock {\em Machine learning}, 45:5--32, 2001.

\bibitem{buturovic2020novel}
Ljubomir Buturovi{\'c} and Dejan Miljkovi{\'c}.
\newblock A novel method for classification of tabular data using convolutional
  neural networks.
\newblock {\em BioRxiv}, pages 2020--05, 2020.

\bibitem{chattopadhay2018grad}
Aditya Chattopadhay, Anirban Sarkar, Prantik Howlader, and Vineeth~N
  Balasubramanian.
\newblock Grad-cam++: Generalized gradient-based visual explanations for deep
  convolutional networks.
\newblock In {\em 2018 IEEE winter conference on applications of computer
  vision (WACV)}, pages 839--847. IEEE, 2018.

\bibitem{chen2016xgboost}
Tianqi Chen and Carlos Guestrin.
\newblock Xgboost: A scalable tree boosting system.
\newblock In {\em Proceedings of the 22nd acm sigkdd international conference
  on knowledge discovery and data mining}, pages 785--794, 2016.

\bibitem{collins2018deep}
Edo Collins, Radhakrishna Achanta, and Sabine Susstrunk.
\newblock Deep feature factorization for concept discovery.
\newblock In {\em Proceedings of the European Conference on Computer Vision
  (ECCV)}, pages 336--352, 2018.

\bibitem{dorogush2018catboost}
Anna~Veronika Dorogush, Vasily Ershov, and Andrey Gulin.
\newblock Catboost: gradient boosting with categorical features support.
\newblock {\em arXiv preprint arXiv:1810.11363}, 2018.

\bibitem{draelos2020use}
Rachel~Lea Draelos and Lawrence Carin.
\newblock Use hirescam instead of grad-cam for faithful explanations of
  convolutional neural networks.
\newblock {\em arXiv e-prints}, pages arXiv--2011, 2020.

\bibitem{frosst2017distilling}
Nicholas Frosst and Geoffrey Hinton.
\newblock Distilling a neural network into a soft decision tree.
\newblock {\em arXiv preprint arXiv:1711.09784}, 2017.

\bibitem{geurts2006extremely}
Pierre Geurts, Damien Ernst, and Louis Wehenkel.
\newblock Extremely randomized trees.
\newblock {\em Machine learning}, 63:3--42, 2006.

\bibitem{gorishniy2022embeddings}
Yura Gorishniy, Ivan Rubachev, and Artem Babenko.
\newblock On embeddings for numerical features in tabular deep learning.
\newblock {\em arXiv preprint arXiv:2203.05556}, 2022.

\bibitem{grinsztajn2022tree}
L{\'e}o Grinsztajn, Edouard Oyallon, and Ga{\"e}l Varoquaux.
\newblock Why do tree-based models still outperform deep learning on tabular
  data?
\newblock {\em arXiv preprint arXiv:2207.08815}, 2022.

\bibitem{huang2020tabtransformer}
Xin Huang, Ashish Khetan, Milan Cvitkovic, and Zohar Karnin.
\newblock Tabtransformer: Tabular data modeling using contextual embeddings.
\newblock {\em arXiv preprint arXiv:2012.06678}, 2020.

\bibitem{humbird2018deep}
Kelli~D Humbird, J~Luc Peterson, and Ryan~G McClarren.
\newblock Deep neural network initialization with decision trees.
\newblock {\em IEEE transactions on neural networks and learning systems},
  30(5):1286--1295, 2018.

\bibitem{joseph2022gate}
Manu Joseph and Harsh Raj.
\newblock Gate: Gated additive tree ensemble for tabular classification and
  regression.
\newblock {\em arXiv preprint arXiv:2207.08548}, 2022.

\bibitem{kadra2021well}
Arlind Kadra, Marius Lindauer, Frank Hutter, and Josif Grabocka.
\newblock Well-tuned simple nets excel on tabular datasets.
\newblock {\em Advances in neural information processing systems},
  34:23928--23941, 2021.

\bibitem{ke2017lightgbm}
Guolin Ke, Qi Meng, Thomas Finley, Taifeng Wang, Wei Chen, Weidong Ma, Qiwei
  Ye, and Tie-Yan Liu.
\newblock Lightgbm: A highly efficient gradient boosting decision tree.
\newblock {\em Advances in neural information processing systems}, 30, 2017.

\bibitem{kontschieder2015deep}
Peter Kontschieder, Madalina Fiterau, Antonio Criminisi, and Samuel~Rota Bulo.
\newblock Deep neural decision forests.
\newblock In {\em Proceedings of the IEEE international conference on computer
  vision}, pages 1467--1475, 2015.

\bibitem{kossen2021self}
Jannik Kossen, Neil Band, Clare Lyle, Aidan~N Gomez, Thomas Rainforth, and
  Yarin Gal.
\newblock Self-attention between datapoints: Going beyond individual
  input-output pairs in deep learning.
\newblock {\em Advances in Neural Information Processing Systems},
  34:28742--28756, 2021.

\bibitem{levin2022transfer}
Roman Levin, Valeriia Cherepanova, Avi Schwarzschild, Arpit Bansal, C~Bayan
  Bruss, Tom Goldstein, Andrew~Gordon Wilson, and Micah Goldblum.
\newblock Transfer learning with deep tabular models.
\newblock {\em arXiv preprint arXiv:2206.15306}, 2022.

\bibitem{lundberg2017unified}
Scott~M Lundberg and Su-In Lee.
\newblock A unified approach to interpreting model predictions.
\newblock {\em Advances in neural information processing systems}, 30, 2017.

\bibitem{mcgill2017deciding}
Mason McGill and Pietro Perona.
\newblock Deciding how to decide: Dynamic routing in artificial neural
  networks.
\newblock In {\em International Conference on Machine Learning}, pages
  2363--2372. PMLR, 2017.

\bibitem{muhammad2020eigen}
Mohammed~Bany Muhammad and Mohammed Yeasin.
\newblock Eigen-cam: Class activation map using principal components.
\newblock In {\em 2020 International Joint Conference on Neural Networks
  (IJCNN)}, pages 1--7. IEEE, 2020.

\bibitem{mullapudi2018hydranets}
Ravi~Teja Mullapudi, William~R Mark, Noam Shazeer, and Kayvon Fatahalian.
\newblock Hydranets: Specialized dynamic architectures for efficient inference.
\newblock In {\em Proceedings of the IEEE Conference on Computer Vision and
  Pattern Recognition}, pages 8080--8089, 2018.

\bibitem{murdock2016blockout}
Calvin Murdock, Zhen Li, Howard Zhou, and Tom Duerig.
\newblock Blockout: Dynamic model selection for hierarchical deep networks.
\newblock In {\em Proceedings of the IEEE conference on computer vision and
  pattern recognition}, pages 2583--2591, 2016.

\bibitem{murthy2016deep}
Venkatesh~N Murthy, Vivek Singh, Terrence Chen, R Manmatha, and Dorin
  Comaniciu.
\newblock Deep decision network for multi-class image classification.
\newblock In {\em Proceedings of the IEEE conference on computer vision and
  pattern recognition}, pages 2240--2248, 2016.

\bibitem{nguyen2020towards}
Duy~T Nguyen, Kathryn~E Kasmarik, and Hussein~A Abbass.
\newblock Towards interpretable anns: An exact transformation to multi-class
  multivariate decision trees.
\newblock {\em arXiv preprint arXiv:2003.04675}, 2020.

\bibitem{popov2019neural}
Sergei Popov, Stanislav Morozov, and Artem Babenko.
\newblock Neural oblivious decision ensembles for deep learning on tabular
  data.
\newblock {\em arXiv preprint arXiv:1909.06312}, 2019.

\bibitem{radenovic2022neural}
Filip Radenovic, Abhimanyu Dubey, and Dhruv Mahajan.
\newblock Neural basis models for interpretability.
\newblock {\em arXiv preprint arXiv:2205.14120}, 2022.

\bibitem{redmon2017yolo9000}
Joseph Redmon and Ali Farhadi.
\newblock Yolo9000: better, faster, stronger.
\newblock In {\em Proceedings of the IEEE conference on computer vision and
  pattern recognition}, pages 7263--7271, 2017.

\bibitem{ribeiro2016should}
Marco~Tulio Ribeiro, Sameer Singh, and Carlos Guestrin.
\newblock " why should i trust you?" explaining the predictions of any
  classifier.
\newblock In {\em Proceedings of the 22nd ACM SIGKDD international conference
  on knowledge discovery and data mining}, pages 1135--1144, 2016.

\bibitem{roy2016monocular}
Anirban Roy and Sinisa Todorovic.
\newblock Monocular depth estimation using neural regression forest.
\newblock In {\em Proceedings of the IEEE conference on computer vision and
  pattern recognition}, pages 5506--5514, 2016.

\bibitem{rubachev2022revisiting}
Ivan Rubachev, Artem Alekberov, Yury Gorishniy, and Artem Babenko.
\newblock Revisiting pretraining objectives for tabular deep learning.
\newblock {\em arXiv preprint arXiv:2207.03208}, 2022.

\bibitem{selvaraju2017grad}
Ramprasaath~R Selvaraju, Michael Cogswell, Abhishek Das, Ramakrishna Vedantam,
  Devi Parikh, and Dhruv Batra.
\newblock Grad-cam: Visual explanations from deep networks via gradient-based
  localization.
\newblock In {\em Proceedings of the IEEE international conference on computer
  vision}, pages 618--626, 2017.

\bibitem{sethi1990entropy}
Ishwar~Krishnan Sethi.
\newblock Entropy nets: from decision trees to neural networks.
\newblock {\em Proceedings of the IEEE}, 78(10):1605--1613, 1990.

\bibitem{shwartz2022tabular}
Ravid Shwartz-Ziv and Amitai Armon.
\newblock Tabular data: Deep learning is not all you need.
\newblock {\em Information Fusion}, 81:84--90, 2022.

\bibitem{simonyan2013deep}
Karen Simonyan, Andrea Vedaldi, and Andrew Zisserman.
\newblock Deep inside convolutional networks: Visualising image classification
  models and saliency maps.
\newblock {\em arXiv preprint arXiv:1312.6034}, 2013.

\bibitem{somepalli2021saint}
Gowthami Somepalli, Micah Goldblum, Avi Schwarzschild, C~Bayan Bruss, and Tom
  Goldstein.
\newblock Saint: Improved neural networks for tabular data via row attention
  and contrastive pre-training.
\newblock {\em arXiv preprint arXiv:2106.01342}, 2021.

\bibitem{sudjianto2020unwrapping}
Agus Sudjianto, William Knauth, Rahul Singh, Zebin Yang, and Aijun Zhang.
\newblock Unwrapping the black box of deep relu networks: interpretability,
  diagnostics, and simplification.
\newblock {\em arXiv preprint arXiv:2011.04041}, 2020.

\bibitem{sun2019supertml}
Baohua Sun, Lin Yang, Wenhan Zhang, Michael Lin, Patrick Dong, Charles Young,
  and Jason Dong.
\newblock Supertml: Two-dimensional word embedding for the precognition on
  structured tabular data.
\newblock In {\em Proceedings of the IEEE/CVF Conference on Computer Vision and
  Pattern Recognition Workshops}, pages 0--0, 2019.

\bibitem{vanschoren2014openml}
Joaquin Vanschoren, Jan~N Van~Rijn, Bernd Bischl, and Luis Torgo.
\newblock Openml: networked science in machine learning.
\newblock {\em ACM SIGKDD Explorations Newsletter}, 15(2):49--60, 2014.

\bibitem{veit2018convolutional}
Andreas Veit and Serge Belongie.
\newblock Convolutional networks with adaptive inference graphs.
\newblock In {\em Proceedings of the European Conference on Computer Vision
  (ECCV)}, pages 3--18, 2018.

\bibitem{wan2020nbdt}
Alvin Wan, Lisa Dunlap, Daniel Ho, Jihan Yin, Scott Lee, Henry Jin, Suzanne
  Petryk, Sarah~Adel Bargal, and Joseph~E Gonzalez.
\newblock Nbdt: neural-backed decision trees.
\newblock {\em arXiv preprint arXiv:2004.00221}, 2020.

\bibitem{wu2018beyond}
Mike Wu, Michael Hughes, Sonali Parbhoo, Maurizio Zazzi, Volker Roth, and
  Finale Doshi-Velez.
\newblock Beyond sparsity: Tree regularization of deep models for
  interpretability.
\newblock In {\em Proceedings of the AAAI conference on artificial
  intelligence}, volume~32, 2018.

\bibitem{yang2018deep}
Yongxin Yang, Irene~Garcia Morillo, and Timothy~M Hospedales.
\newblock Deep neural decision trees.
\newblock {\em arXiv preprint arXiv:1806.06988}, 2018.

\bibitem{zeiler2014visualizing}
Matthew~D Zeiler and Rob Fergus.
\newblock Visualizing and understanding convolutional networks.
\newblock In {\em European conference on computer vision}, pages 818--833.
  Springer, 2014.

\bibitem{zhang2018top}
Jianming Zhang, Sarah~Adel Bargal, Zhe Lin, Jonathan Brandt, Xiaohui Shen, and
  Stan Sclaroff.
\newblock Top-down neural attention by excitation backprop.
\newblock {\em International Journal of Computer Vision}, 126(10):1084--1102,
  2018.

\bibitem{zhang2018tropical}
Liwen Zhang, Gregory Naitzat, and Lek-Heng Lim.
\newblock Tropical geometry of deep neural networks.
\newblock In {\em International Conference on Machine Learning}, pages
  5824--5832. PMLR, 2018.

\bibitem{zhou2016learning}
Bolei Zhou, Aditya Khosla, Agata Lapedriza, Aude Oliva, and Antonio Torralba.
\newblock Learning deep features for discriminative localization.
\newblock In {\em Proceedings of the IEEE conference on computer vision and
  pattern recognition}, pages 2921--2929, 2016.

\bibitem{zhu2021converting}
Yitan Zhu, Thomas Brettin, Fangfang Xia, Alexander Partin, Maulik Shukla,
  Hyunseung Yoo, Yvonne~A Evrard, James~H Doroshow, and Rick~L Stevens.
\newblock Converting tabular data into images for deep learning with
  convolutional neural networks.
\newblock {\em Scientific reports}, 11(1):11325, 2021.

\end{thebibliography}
}

\end{document}